# The Convergent Ethics of AI? Analyzing Moral Foundation Priorities in Large Language Models with a Multi-Framework Approach

Chad Coleman
New York University
cjc652@nyu.edu

W. Russell Neuman
New York University
wrn210@nyu.edu

Ali Dasdan
DropBox
alidasdan@gmail.com

Safinah Ali
New York University
sa1940@nyu.edu

Manan Shah
New York University
ms10537@nyu.edu

## Abstract

As large language models (LLMs) are increasingly deployed in consequential decision-making contexts, systematically assessing their ethical reasoning capabilities becomes a critical imperative. This paper introduces the Priorities in Reasoning and Intrinsic Moral Evaluation (PRIME) framework—a comprehensive methodology for analyzing moral priorities across foundational ethical dimensions including consequentialist-deontological reasoning, moral foundations theory, and Kohlberg's developmental stages. We apply this framework to six leading LLMs through a dual-protocol approach combining direct questioning and response analysis to established ethical dilemmas. Our analysis reveals striking patterns of convergence: all evaluated models demonstrate strong prioritization of care/harm and fairness/cheating foundations while consistently underweighting authority, loyalty, and sanctity dimensions. Through detailed examination of confidence metrics, response reluctance patterns, and reasoning consistency, we establish that contemporary LLMs (1) produce decisive ethical judgments, (2) demonstrate notable cross-model alignment in moral decision-making, and (3) generally correspond with empirically established human moral preferences. This research contributes a scalable, extensible methodology for ethical benchmarking while highlighting both the promising capabilities and systematic limitations in current AI moral reasoning architectures—insights critical for responsible development as these systems assume increasingly significant societal roles.

The rapid evolution of generative large language models (LLMs) has brought the alignment issue to the forefront of AI ethics discussions - specifically, whether these models are appropriately aligned with human values (Bostrom, 2014; Tegmark 2017; Russell 2019; Kosinski, 2024). As these powerful models are increasingly integrated into decision-making processes across various societal domains (Salazar, A., & Kunc, M., 2025), understanding whether and how their operational logic aligns with fundamental human values becomes not just an academic question, but a critical societal imperative. Three questions arise. 1) Will these models respond with clear-cut decisions when confronted with moral dilemmas? 2) Are the responses of different LLMs convergent? 3) Do patterns of LLMs' moral choices align meaningfully with human values broadly defined? In this paper we will present an analytical framework and findings to address the first two questions, and a preliminary exploratory analysis of the third. We will make the case that the answers to these questions are: yes, yes and yes. There are caveats and exceptions, of course, but the broad pattern, we believe, is clear. Our methodology permits us to explore not just what choices they make, but the reasoning chain of thought that leads to those decisions.

The tradition of moral philosophy typically does not provide definitive "right" answers to specific moral dilemmas and complex scenarios (Grassian 1992; Joyce 2006; Jonsen & Toulmin, 1988). Instead, this literature provides frameworks and principles to help analyze different perspectives on situational challenges (Neuman, Coleman & Shah 2025; Bickley & Torgler, 2023). Therefore, while the ground truth approach to benchmarking and comparative model performance metrics is prominently used in the technical literature (Reuel et al. 2024), it proves inadequate for understanding ethical reasoning. What is needed instead is a systematic framework for analyzing how these models explain and justify their ethical decisions.

We propose a novel analytical framework that leverages three established typologies from moral philosophy, comparative ethics, and behavioral psychology to examine LLM ethical reasoning. One key lens, Moral Foundations Theory (Haidt, 2012; Graham et al., 2009), allows us to probe beyond simple judgments, examining the relative weight given to different moral intuitions such as Care and Fairness versus Loyalty, Authority, and Sanctity. Recognizing established patterns in human moral psychology, such as the tendency for different societal groups (often characterized along liberal-conservative lines) to prioritize these foundations differently (Graham et al., 2009; Haidt, 2012), opens avenues for investigating whether similar patterns or biases manifest in LLM reasoning. This overall approach resonates with what Gunning (2019) calls 'explainable AI' while providing a rigorous structure for comparative analysis across different ethical dimensions (Kohlberg, 1981; Shneiderman, 2022). This framework focuses not on evaluating the correctness of ethical decisions, but on systematically analyzing how these systems explain and justify their choices through both direct self-description and responses to classic moral dilemmas. Crucially, this multi-framework approach is presented not only to analyze the specific models in this study but also to offer a robust, extensible methodology for researchers to systematically evaluate the ethical reasoning capabilities of current and future AI systems.

Our methodology comprises two distinct procedures: 1) Direct questioning of the model about its ethical decision-making processes using standardized prompts and typological frameworks, and 2) Analysis of the model's responses to classic moral dilemmas, followed by structured assessment of its reasoning explanations. The framework employs three complementary analytical lenses: the consequentialist-deontological dichotomy (Alexander & Moore 2024), Moral Foundations Theory (Haidt & Craig 2004), and Kohlberg's Stages of Moral Development (Kohlberg 1964, 1981).

To test and demonstrate this framework, we examine six widely-used generative AI tools: OpenAI's GPT4o, Meta's LLaMA 3.1, Perplexity, Anthropic's Claude 3.5 Sonnet, Google's Gemini and Mistral's 7B. While our findings are necessarily limited to the current generation of transformer models, the framework itself is designed to be extensible and applicable to future AI systems.

## Problem Statement

As artificial intelligence systems increasingly engage in moral decision-making, understanding their moral reasoning patterns and capabilities becomes crucial for responsible AI development and deployment. Current literature lacks a comprehensive analysis of how different language models approach moral dilemmas across various ethical frameworks, particularly in terms of their moral foundation preferences, decision confidence, and reasoning patterns. While existing research has examined AI decision-making in specific ethical scenarios, there remains a significant gap in our understanding of how different language models systematically process and respond to moral challenges across multiple ethical dimensions.

The complexity of this challenge is compounded by the interaction between different moral foundations in AI reasoning, the variation in confidence levels across different types of ethical decisions, and the potential biases in how these systems prioritize certain moral considerations over others. Understanding these patterns and variations is essential for developing more balanced and ethically sound AI systems, as well as for identifying potential limitations or biases in current approaches to AI moral reasoning.

## Research Questions

- How do contemporary language models vary in their approach to moral reasoning across different ethical foundations?
- What patterns emerge in their decision-making processes, confidence levels, and moral foundation preferences when confronted with complex ethical dilemmas?

These questions draw upon prior work examining AI ethical frameworks (Etzioni & Etzioni, 2017; Wallach & Allen, 2009) while focusing specifically on the reasoning patterns that emerge across different moral foundations and specifically examine the interplay between different moral foundations in AI reasoning, the relationship between decision confidence and reasoning sophistication, and the systematic patterns in how different models prioritize and process various ethical considerations.

# Implementing the Framework: Core Components and Assessment Protocol

The systematic analysis of ethical reasoning in LLMs requires careful attention to both the elicitation of responses and their subsequent analysis. Our framework, indicated in Figure 1, establishes specific protocols that maintain flexibility while accommodating the evolving capabilities of these systems (Hagendorff & Danks, 2023; Prem, 2023). The framework includes two key elements: Response Elicitation, and Analytical Assessment, that are outlined below.

The Priorities in Reasoning and Intrinsic Moral Evaluation (PRIME) framework employs dual elicitation strategies designed to reveal different aspects of LLM ethical reasoning processes. The first component focuses on self-descriptive ethical analysis through direct questioning about moral decision-making processes. This includes queries about general ethical reasoning, specific questions related to established moral frameworks, and meta-cognitive prompts about the system's understanding of its own reasoning (Butlin et al., 2023; Atari et al., 2023).

The second component examines applied reasoning through established ethical dilemmas. These scenarios include the classic Trolley Problem to explore consequentialist versus deontological thinking (Greene, 2023; Thomson, 1976), the Heinz Dilemma to examine moral development stages (Kohlberg, 1981; Rest, 1979), and the Lifeboat Scenario to assess complex multi-party ethical calculations (Brzozowski, 2003; Hardin, 1974). Game theoretic scenarios such as the Prisoner's Dilemma and Dictator's Game provide additional insights into strategic ethical thinking (Peterson, 2015; Axelrod, 1984).

The selection of game theory scenarios such as the Prisoner's Dilemma and Dictator's Game for this study represents a strategic methodological choice that enhances the analytical framework's depth and ecological validity. Game theory scenarios were incorporated because they offer structured ethical decision-making contexts that require balancing self-interest against collective welfare, thus revealing how language models navigate complex social dilemmas with competing moral imperatives. Unlike more abstract ethical questions, these scenarios provide quantifiable outcomes and clear decision points that allow researchers to systematically compare different AI systems' ethical reasoning patterns, strategic thinking capabilities, and moral foundation preferences. Furthermore, game theory dilemmas have been extensively studied in human moral psychology (as evidenced by references to Peterson, 2015 and Axelrod, 1984), offering established benchmarks against which AI moral reasoning can be compared. By including these scenarios alongside classic moral dilemmas like the Trolley Problem and Heinz Dilemma, we created a comprehensive assessment protocol capable of examining how language models balance different ethical considerations across varied contexts, thus addressing the research questions about variation in moral reasoning approaches and patterns in decision-making processes.

Figure 1: Flow-chart of proposed PRIME framework for the assessment of LLM responses

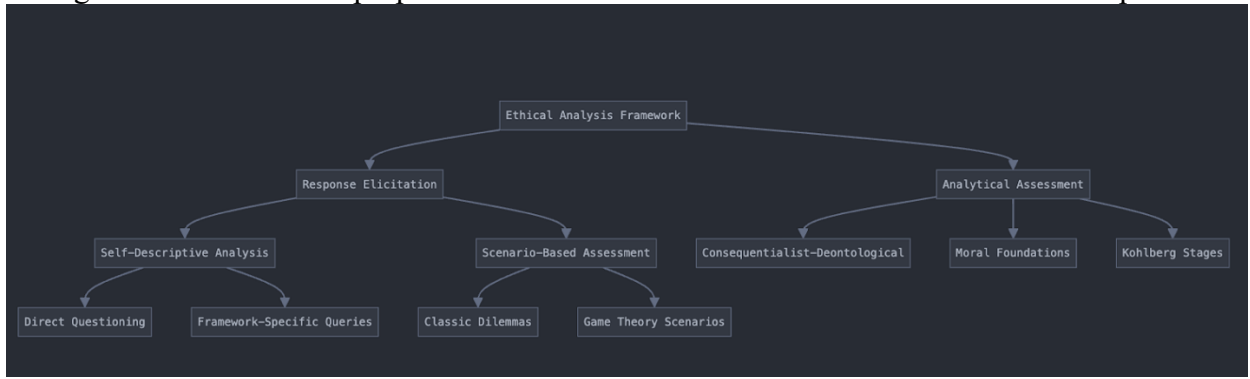

PRIME applies three complementary analytical lenses to examine responses. The first examines the balance between consequentialist and deontological reasoning, as established in classical moral philosophy (Alexander & Moore, 2024). This analysis reveals whether systems prioritize outcome-based reasoning, rule-based principles, or employ hybrid approaches that combine both perspectives.

The second analytical lens employs Moral Foundations Theory to assess responses through fundamental moral intuitions (Haidt & Craig, 2004). This includes examination of how systems weigh considerations of care and harm prevention, fairness and reciprocity, loyalty to collective interests, respect for legitimate authority, and adherence to higher principles. The relative emphasis on these foundations provides insight into how LLMs prioritize different moral considerations.

The third dimension maps responses to Kohlberg's stages of moral development, ranging from pre-conventional focus on punishment and self-interest, through conventional emphasis on social norms, to post-conventional consideration of universal principles (Kohlberg, 1981). This mapping helps reveal the sophistication and consistency of ethical reasoning across different scenarios.

Following Reuel et al. (2024), we implemented multiple independent analysts to ensure reliable coding and pattern identification across the different models. The assessment process begins with comprehensive documentation of initial responses, including all clarifications and elaborations. Multiple independent analysts apply the three analytical frameworks to ensure reliable coding and pattern identification. This process enables both within-model analysis across scenarios and cross-model comparison of reasoning patterns.

Quality control measures include standardized prompt formulation, regular calibration of analysis procedures, and systematic investigation of anomalous responses. The framework emphasizes documentation of both patterns and edge cases, maintaining methodological rigor while remaining open to unexpected insights about LLM ethical reasoning (Bickley & Torgler, 2023).

# Methodology

This study employed a mixed-methods approach (Reuel et al., 2024) to analyze moral reasoning patterns in large language models. The research design incorporated both quantitative analysis of moral foundation scores and qualitative assessment of response patterns, utilizing a comparative framework to examine six different LLMs: OpenAI's GPT4o, Meta's LLaMA 3.1, Perplexity, Anthropic's Claude 3.5 Sonnet, Google's Gemini and Mistral's 7B. For simplicity, these models will be referred to as ChatGPT, LLaMA, Perplexity, Claude, Gemini, and Mistral, respectively. These models were selected based on their widespread use and varying architectural approaches, representing a diverse range of training methodologies and capabilities, which allowed for comprehensive comparison of moral reasoning patterns across different AI systems.

The data collection process consisted of three main phases. First, a standardized set of moral dilemmas was presented to each LLM, designed to probe different moral foundations based on Moral Foundations Theory (Thoma & Dong, 2014). Each model received identical scenarios to ensure consistency in comparison, and multiple responses were collected for each dilemma to assess response stability. Second, three sequential responses were gathered for each dilemma, recorded verbatim, including any expressions of uncertainty or qualification. Time stamps were logged for each response to analyze decision-making patterns, and all interactions were conducted using standardized prompts to maintain consistency. Third, models were prompted to provide confidence scores (0-100) for their decisions, with confidence assessments collected immediately following each moral judgment. Additional prompts were used to elicit explanations for confidence ratings.

The study utilized several key instruments and measures. The six-dimensional moral foundations framework included care/harm, fairness/cheating, loyalty/betrayal, authority/subversion, sanctity/degradation, and liberty/oppression, with each dimension scored on a 1-5 scale based on standardized criteria. The Kohlberg Moral Reasoning Scale (Colby & Kohlberg, 1987) was applied to assess the sophistication of moral reasoning, with scores ranging from 1-6, corresponding to Kohlberg's stages of moral development. Two independent raters evaluated responses using standardized rubrics. Additionally, a custom metric was developed to quantify decision-making hesitation, incorporating factors such as number of qualifications in responses, expression of uncertainty, time taken to reach final decisions, and changes in position across multiple responses.

The data analysis phase incorporated both quantitative and qualitative methods. Quantitative analysis included correlation analysis, with Pearson correlation coefficients calculated between moral foundation scores and statistical significance tested at $p < 0.05$. Confidence score analysis utilized box plots to compare distributions and regression analysis to examine relationships with Kohlberg scores. Response pattern analysis employed stacked bar charts to visualize decision timing, with reluctance scores computed and compared across models. Qualitative analysis involved text analysis using Natural Language Processing (NLP) techniques, including word frequency analysis, sentiment analysis, and topic modeling to identify key themes (Schramowski et al., 2022). Moral foundation mapping utilized radar charts to visualize moral foundation profiles, complemented by content analysis to identify foundation-specific language.

To ensure validity and reliability, several measures were implemented. Two independent raters scored responses, with Cohen's kappa calculated to assess agreement and discrepancies resolved through consensus discussion. Cross-validation was performed using multiple scenarios to test consistency of responses, with patterns verified across multiple testing sessions. The analysis utilized a Python-based pipeline for data processing, with statistical analyses performed using standard scientific computing libraries and visualization tools including matplotlib and seaborn. Custom scripts were developed for moral foundation scoring.

The research team carefully considered ethical implications and potential biases in scenario design and interpretation. While working with AI systems rather than human subjects minimized traditional ethical concerns, attention was paid to fairness in scenario presentation, transparency in scoring criteria, potential biases in training data, and implications of findings for AI deployment. Several methodological controls were implemented to address potential limitations, including standardized prompts to minimize variation in question presentation, multiple testing sessions to account for response variability, balanced representation of different moral foundations, and cross-validation of scoring methods. This comprehensive methodology was designed to provide rigorous analysis of moral reasoning in LLMs while maintaining reproducibility and reliability in the research process.

## Findings

This section presents a comprehensive analysis of moral reasoning patterns across various language models (LLMs). We investigated the complex interplay between different moral foundations, confidence levels, response reluctance, and linguistic patterns exhibited by six prominent LLMs: ChatGPT, Claude, Gemini, LLaMA, Mistral, and Perplexity. Our multifaceted examination employed correlation analysis, confidence assessment, response pattern analysis, moral foundation profiling, and text analysis to uncover systematic differences in how these models approach ethical decision-making. These findings provide valuable insights into the moral reasoning capabilities of contemporary LLMs, revealing both consistencies across models and notable variations that reflect their underlying architectural and training differences. The following subsections detail our specific observations across each analytical dimension.

Figure 2: Correlation matrix of moral foundation scores across LLM responses.

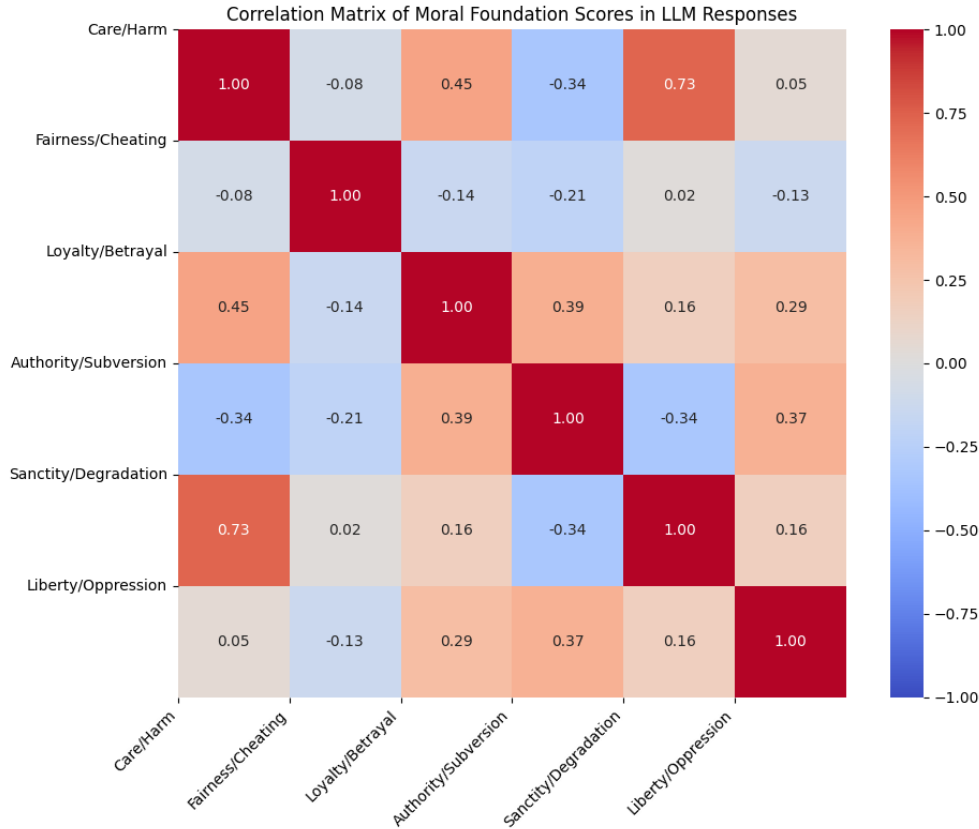

The correlation matrix in Figure 2 reveals several notable relationships between different moral foundation scores in LLM responses. The strongest positive correlation (r = 0.73) was observed between care/harm and sanctity/degradation scores, suggesting these moral foundations often co-occur in LLM outputs. Loyalty/betrayal showed moderate positive correlations with care/harm (r = 0.45) and authority/subversion (r = 0.39), indicating some alignment between these moral dimensions.

Interestingly, authority/subversion demonstrated a negative correlation with care/harm (r = -0.34) and sanctity/degradation (r = -0.34), suggesting potential tensions between these moral foundations in LLM responses. The fairness/cheating dimension showed weak negative correlations with most other foundations, with the strongest negative relationship being with authority/subversion (r = -0.21). Liberty/oppression scores showed relatively weak correlations with most other foundations, except for a moderate positive correlation with authority/subversion (r = 0.37). This suggests that liberty considerations in LLM responses may operate somewhat independently from other moral foundations.

These patterns indicate that moral foundations in LLM responses are not uniformly aligned but rather show complex interactions, with some foundations showing strong positive associations while others demonstrate negative or negligible relationships. This complexity mirrors findings from human moral psychology studies (Atari et al., 2023), suggesting LLMs may be capturing some of the nuanced interactions between moral intuitions observed in human reasoning and mimics the nuanced nature of moral reasoning these systems attempt to engage in.

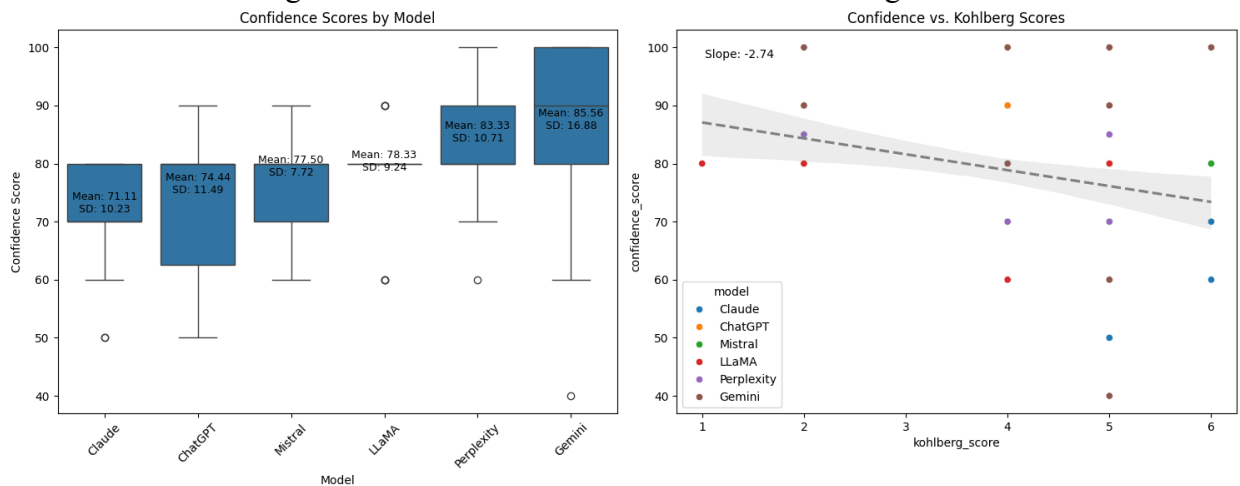

Figure 3: Distribution of confidence and Kohlberg scores

Figure 3 presents two complementary visualizations examining confidence scores across different language models and their relationship with Kohlberg moral reasoning scores. The first visualization is a box plot comparing confidence scores across six different language models, while the second is a scatter plot depicting the relationship between confidence scores and Kohlberg moral reasoning scores.

Analysis of confidence scores revealed significant variations across the six language models examined (see Figure 3). Gemini demonstrated the highest mean confidence (M = 85.56, SD = 16.88), followed by Perplexity (M = 83.33, SD = 10.71). LLaMA (M = 78.33, SD = 9.24), Mistral (M = 77.50, SD = 7.72), and ChatGPT (M = 74.44, SD = 11.49) showed moderate confidence levels. Claude exhibited the lowest average confidence (M = 71.11, SD = 10.23).

The box plot reveals notable variations in confidence distributions. Gemini and Perplexity demonstrate higher median confidence scores (approximately 90 and 85, respectively) compared to other models. Claude, ChatGPT, and Mistral exhibit similar median confidence scores in the 70-80 range, with comparable interquartile ranges and more symmetric distributions. LLaMA shows a relatively narrow distribution centered around 80, though it includes several outliers at both extremes. All models display outliers, particularly in lower ranges, suggesting occasional instances of marked uncertainty in moral reasoning.

Post-hoc analyses using Tukey's HSD test identified three statistically significant differences in confidence expression between models. Gemini demonstrated significantly higher confidence than both Claude (mean difference = 14.44, p = .003) and ChatGPT (mean difference = 11.11, p = .048). Additionally, Perplexity showed significantly higher confidence than Claude (mean difference = 12.22, p = .021). No other pairwise comparisons reached statistical significance (p > .05).

The scatter plot examining the relationship between confidence scores and Kohlberg moral reasoning scores reveals a slight negative correlation, as indicated by the downward-sloping regression line (slope = -2.74, intercept = 89.81). This suggests that models tend to express somewhat lower confidence when engaging with more complex moral reasoning scenarios (higher Kohlberg scores). The confidence scores range from approximately 40 to 100 across all Kohlberg levels (1-6). Notably, there is considerable variation in confidence scores at each Kohlberg level, indicating that the relationship between reasoning complexity and confidence is not strongly deterministic.

Analysis of Kohlberg scores revealed differences in moral reasoning sophistication across models. Claude demonstrated the highest average Kohlberg score (M = 4.56, SD = 1.10), while LLaMA exhibited the lowest (M = 3.72, SD = 1.45). Notably, models with the highest confidence scores (Gemini and Perplexity) did not necessarily demonstrate the most sophisticated moral reasoning, suggesting a potential disconnect between confidence and reasoning quality.

The data reveals interesting patterns in how different models express confidence in their moral decisions. While some models (particularly Gemini and Perplexity) consistently express high confidence, others maintain more moderate confidence levels. The negative correlation between confidence and Kohlberg scores suggests appropriate calibration of confidence relative to problem complexity, though this relationship is modest. This pattern aligns with research on human moral judgment (Rathi & Kumar, 2020), where increased awareness of moral complexity often corresponds with greater epistemic humility.

These findings have important implications for AI system development and deployment in contexts requiring moral reasoning, particularly regarding appropriate uncertainty expression. The variation in confidence patterns across models suggests that architectural and training differences may significantly influence how models assess their own certainty in moral reasoning tasks. Further research is needed to understand the factors driving these confidence patterns and their relationship to performance accuracy.

Figure 4: Frequency of response reluctance across LLMs

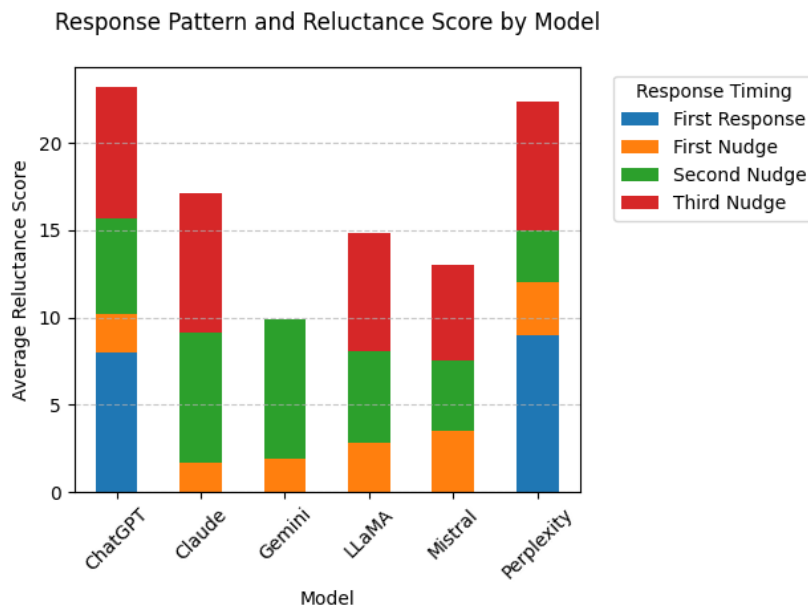

Figure 4 shows a comparative analysis of response patterns and reluctance scores across six different language models (ChatGPT, Claude, Gemini, LLaMA, Mistral, and Perplexity), offering insights into how these models approach complex moral decision-making. The visualization combines two key metrics: the timing of final decisions (indicated by color-coded segments representing first, second, or third responses) and the overall reluctance scores for each model.

A notable pattern emerges in the distribution of reluctance scores, with ChatGPT and Perplexity exhibiting the highest overall reluctance scores (approximately 23 points each), suggesting these models demonstrate greater hesitation or careful deliberation in their decision-making process. In contrast, Mistral and LLaMA show lower overall reluctance scores (around 13-15 points), potentially indicating more direct or decisive response patterns. Claude and Gemini fall in the middle range, with reluctance scores of approximately 17 and 10 points respectively.

Table 1: Model Reluctance Scores

| Model | First | One Nudge | Two Nudges | Three Nudges |
|---|---|---|---|---|
| ChatGPT | 8 | 2.2 | 5.5 | 7.5 |
| Claude |  | 1.64 | 7.5 | 8 |
| Gemini |  | 1.88 | 8 |  |
| LLaMA |  | 2.83 | 5.25 | 6.75 |
| Mistral |  | 3.5 | 4 | 5.5 |
| Perplexity | 9 | 3 | 3 | 7.36 |

The timing of final decisions reveals interesting variations across models. ChatGPT and Perplexity show substantial first-response components (represented by blue segments), indicating they often reach their final decisions earlier in the interaction. However, they maintain high reluctance scores, suggesting that while they may decide quickly, they express significant consideration in their responses. Conversely, Gemini and Claude demonstrate a more distributed decision-making pattern, with final decisions spread across all three response opportunities (first, second, and third responses), potentially indicating a more iterative decision-making process.

These findings provide valuable insights into the deliberative capabilities of different language models when faced with moral decisions. The variation in reluctance scores and response timing patterns suggests that different models employ distinct approaches to ethical reasoning. Some models, like ChatGPT and Perplexity, combine quick decision-making with high expressed reluctance, while others, like Gemini and Claude, show more distributed decision-making patterns. This diversity in approach raises interesting questions about the optimal balance between decisiveness and careful deliberation in AI systems dealing with moral questions, such as how AI companies choose to create safeguards for LLMs to not appear decisive around moral decision-making prompts, how these safeguarding procedures may differ across LLMs, how these strategies may turn beneficial or harmful in real-world moral decision-making scenarios, where decision-making time is of key importance, and how this deliberation may be bypassed by simple nudging strategies.

The observation about varying safeguarding levels across LLMs raises crucial questions about AI ethical guardrails. The disparate approaches to moral scenario handling indeed suggest differing priorities among AI developers - some emphasizing stronger reluctance mechanisms while others permit more decisiveness. However, the universally successful bypassing of these safeguards through simple nudging techniques exposes a fundamental vulnerability in current systems. This challenges us to reconsider how effective these protections truly are and whether they provide meaningful safety or merely superficial restraint. The balance between deliberative caution and decisive action represents a central tension in AI ethics development - excessive hesitation could render AI unhelpful in time-sensitive situations requiring moral judgment, while insufficient safeguards risk premature or harmful conclusions. This highlights the need for more sophisticated guardrail systems that can maintain appropriate reluctance while resisting manipulation, particularly as these systems become increasingly embedded in consequential decision-making contexts.

A limitation of this analysis is that it does not reveal the nuance or appropriateness of the final decisions made, only the pattern of how and when these decisions were reached. Further research could investigate the relationship between these response patterns and the ethical validity of the final decisions, as well as examine whether higher reluctance scores correlate with more nuanced or ethically sound responses.

A comparative analysis of moral foundation orientations across six different language models (Claude, ChatGPT, Mistral, LLaMA, Perplexity, and Gemini) using radar charts was completed (Figure 5). Each line maps five key moral dimensions: Loyalty, Fairness, Care, Liberty, and Sanctity, with Authority positioned between Loyalty and Sanctity. The radar charts utilize a scale from 1 to 4, radiating outward from the center, to quantify the relative emphasis each model places on these moral foundations.

Figure 5: Model Features Radar Distribution

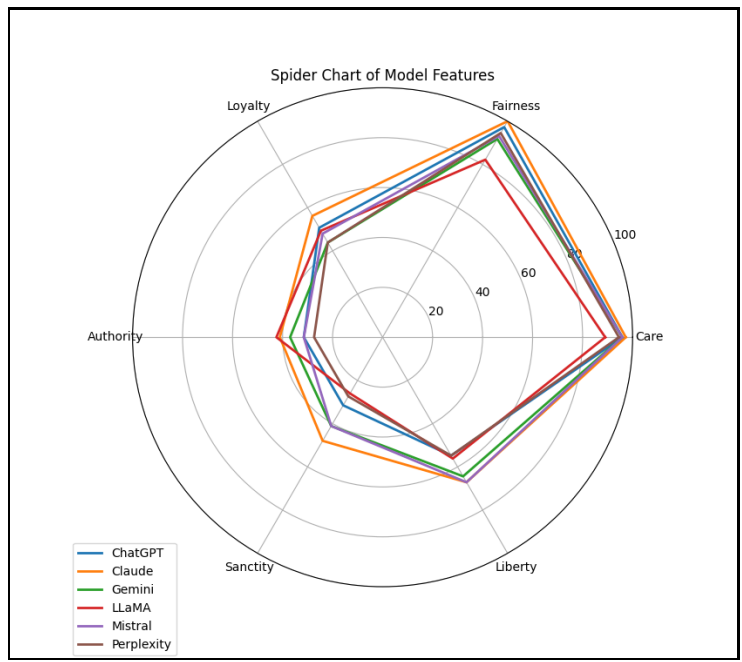

A striking pattern emerges across all six models, showing remarkable consistency in their moral foundation profiles. The most pronounced emphasis appears in the Care and Fairness dimensions, consistently reaching towards the outer rings of the charts. This suggests that these language models have been developed with a strong orientation towards ethical principles centered on harm prevention and equitable treatment. In contrast, the Authority, Loyalty, and Sanctity foundations generally show lower readings.

Table 2: Foundation Score Statistics

|  | Care Mean | Fairness Mean | Loyalty Mean | Authority Mean | Sanctity Mean | Liberty Mean |
|---|---|---|---|---|---|---|
| **ChatGPT** | 3.889 | 3.944 | 2.056 | 1.278 | 1.278 | 2.222 |
| **Claude** | 3.944 | 4.056 | 2.278 | 1.667 | 1.944 | 2.722 |
| **Gemini** | 3.889 | 3.722 | 1.778 | 1.500 | 1.667 | 2.611 |
| **LLaMA** | 3.611 | 3.333 | 2.000 | 1.722 | 1.056 | 2.278 |
| **Mistral** | 3.889 | 3.778 | 1.944 | 1.278 | 1.667 | 2.722 |
| **Perplexity** | 3.833 | 3.833 | 1.778 | 1.111 | 1.111 | 2.222 |

Notably, while the general pattern is similar across models, there are subtle variations in their moral foundation profiles. For instance, ChatGPT and Mistral display nearly identical patterns, with particularly strong emphasis on Care and Fairness, while maintaining moderate levels of other foundations. Claude shows a slightly more balanced distribution across all foundations, though still prioritizing Care and Fairness. The Liberty dimension consistently maintains moderate to high levels across all models, suggesting a balanced consideration of individual freedom within their ethical frameworks.

These findings have important implications for understanding the ethical architectures of current language models. The consistent prioritization of Care and Fairness across different models suggests either an inherent bias in training methodologies or a deliberate design choice reflecting contemporary ethical priorities. The relatively lower emphasis on Authority, Loyalty, and Sanctity might indicate a more individualistic and harm-based moral framework rather than a community or tradition-based ethical system.

One limitation of this visualization is that it captures static moral foundation weightings, while real-world ethical reasoning often requires dynamic balancing of these principles depending on context. Additionally, the quantification of these moral dimensions on a 1-4 scale may oversimplify the complex nature of ethical reasoning. Future research might benefit from examining how these moral foundation profiles manifest in specific ethical decision-making scenarios and whether they remain consistent across different types of moral dilemmas.

Analyzing Kohlberg scores (Figure 6 below) each subplot compares the score (shown in blue) with a specific ethical dimension (shown in orange), demonstrating how moral reasoning sophistication correlates with different aspects of ethical decision-making.

The plots reveal several notable patterns across the tested models (Claude, ChatGPT, Mistral, LLaMA, Perplexity and Gemini). The Kohlberg scores generally maintain a relatively stable range between 3.5 and 4.5 across most models, with some fluctuation. A consistent dip in Kohlberg scores is observed for the LLaMA model across multiple metrics, suggesting potentially less sophisticated moral reasoning capabilities in this model compared to its counterparts.

Particularly interesting relationships emerge in several comparisons. The confidence scores show an inverse relationship with Kohlberg scores, with confidence generally increasing while Kohlberg scores remain stable or slightly decrease, potentially indicating that higher confidence doesn't necessarily correlate with more sophisticated moral reasoning. The care/harm scores maintain a relatively parallel but lower trajectory compared to Kohlberg scores, suggesting a consistent relationship between these two measures of ethical reasoning.

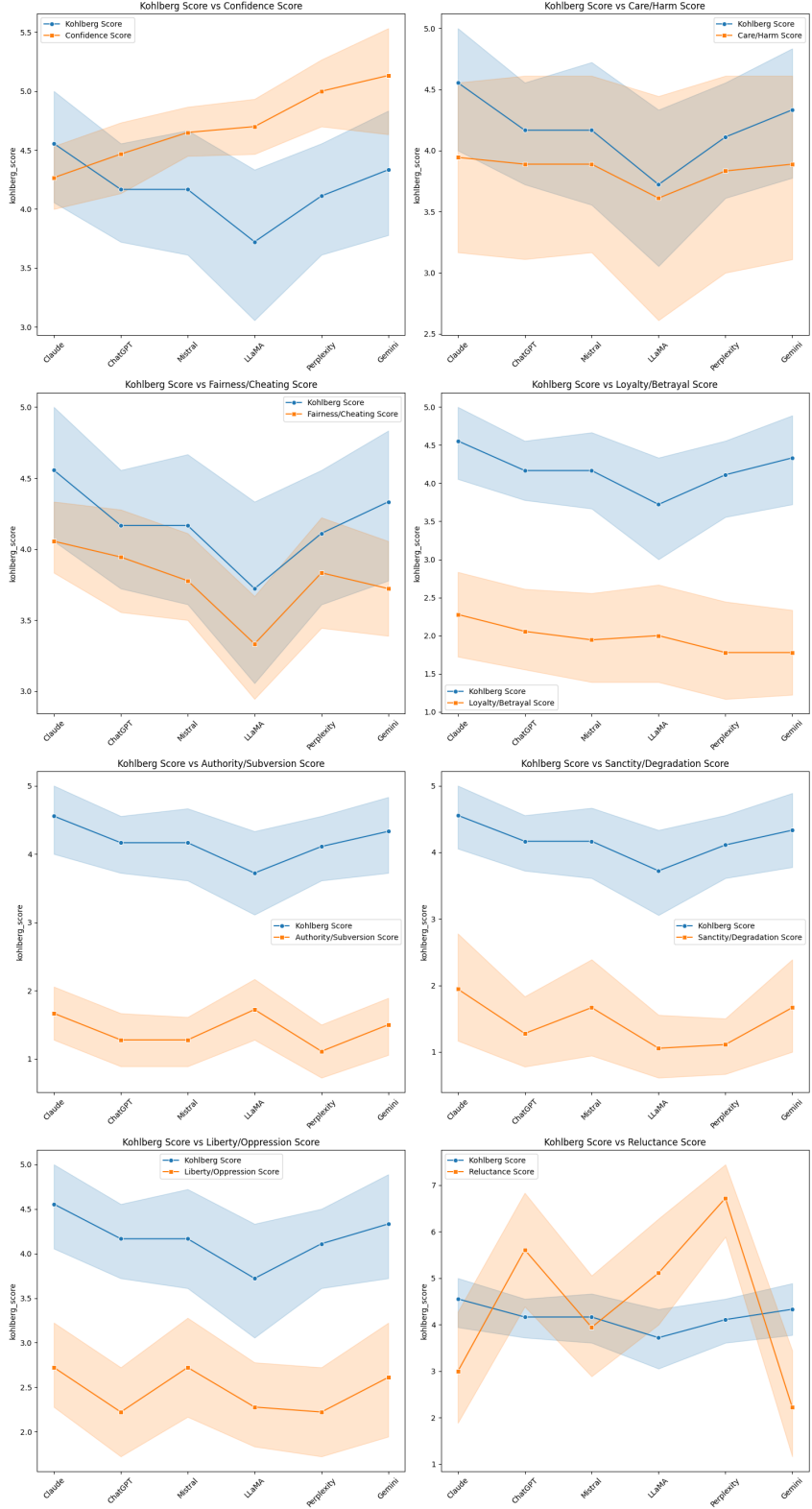

Figure 6: Relationship between Kohlberg Scores and Other Ethical Metrics

The fairness/cheating and loyalty/betrayal comparisons reveal larger gaps between the Kohlberg scores and their respective ethical metrics, with the ethical dimension scores consistently lower than the Kohlberg scores. Kohlberg scores range from 3.7 to 4.5 while fairness/cheating scores range from 3.3 to 4.0, creating a gap of 0.5-0.8 points. For loyalty/betrayal, Kohlberg scores maintain a similar 3.7-4.5 range, but loyalty/betrayal scores are substantially lower at 1.8-2.2, representing a gap of approximately 2.0-2.5 points.

An interesting anomaly appears in the Reluctance Score comparison, which shows the highest variability among all metrics, with dramatic fluctuations and the widest confidence intervals. This suggests that models' hesitation or reluctance in ethical decision-making may be less stable and more context-dependent than other ethical metrics. The shaded areas representing confidence intervals in each plot indicate varying levels of uncertainty across different models and metrics. These intervals generally widen around the LLaMA model's measurements, suggesting greater uncertainty in this model's ethical reasoning capabilities. The consistency of these patterns across multiple ethical dimensions strengthens the reliability of these observations and suggests systematic differences in how different language models approach moral reasoning tasks.

These findings provide valuable insights into the moral reasoning capabilities of different language models while highlighting the complex relationships between various aspects of ethical decision-making. The consistent patterns across multiple ethical dimensions suggest that these relationships are not random but reflect fundamental characteristics of how these models process and respond to moral dilemmas.

The final analysis examines moral reasoning patterns across different ethical foundations through two complementary approaches: topic modeling and sentiment analysis.

Figure 7: Average Moral Foundation Scores by Model

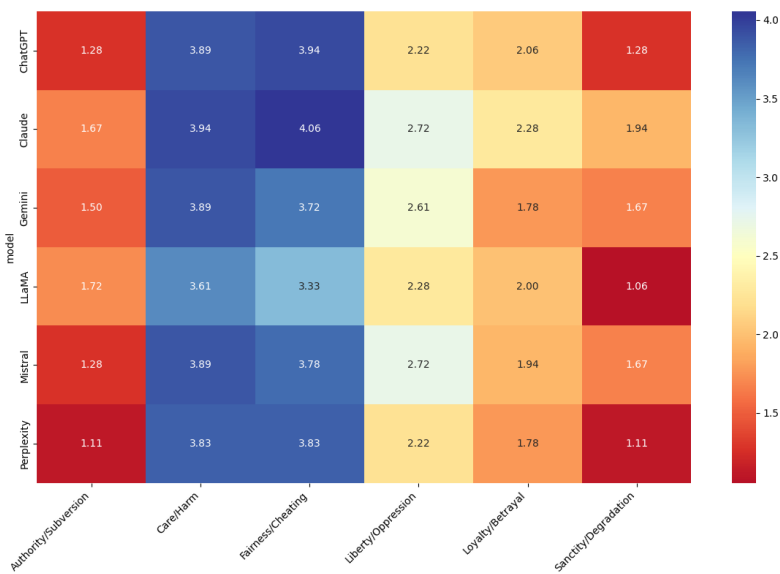

The heatmap (Figure 7) comparing moral foundation scores across different AI models (ChatGPT, Claude, Gemini, LLaMA, Mistral, and Perplexity) reveals several notable patterns. Care/harm and fairness/cheating consistently receive the highest scores across all models (ranging from 3.3 to 4.1), indicating these foundations are most prominently represented in AI moral reasoning. Claude demonstrates particularly high scores in these domains (3.94 for care/harm and 4.06 for fairness/cheating). In contrast, authority/subversion and sanctity/degradation foundations receive notably lower scores (typically between 1.1 and 1.7), suggesting these moral considerations are less emphasized in AI responses.

Figure 8: Average Sentiment Scores Across Moral Foundations

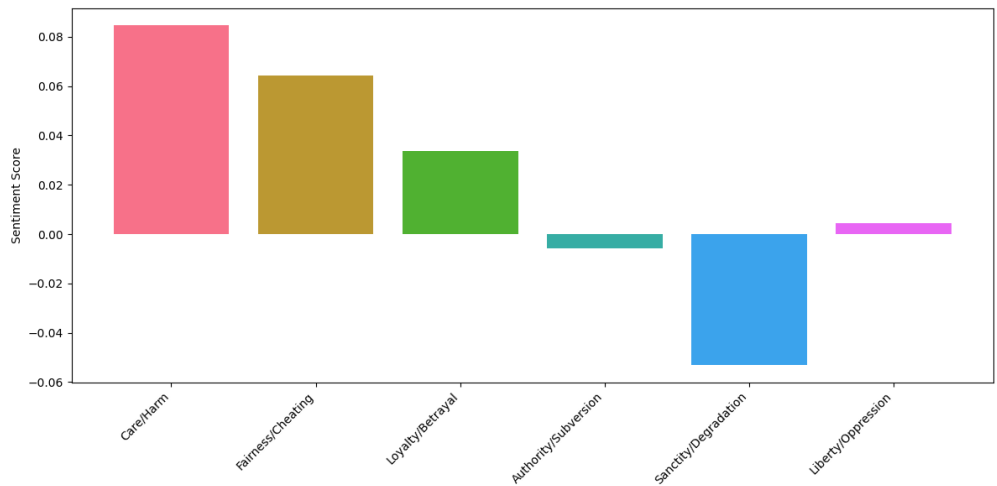

The sentiment analysis bar graph found in Figure 8 provides additional insight into the emotional valence associated with different moral foundations. Care/harm shows the highest positive sentiment score (0.085), followed by fairness/cheating (0.064), indicating these foundations tend to elicit more positive emotional responses. Interestingly, sanctity/degradation displays the most negative sentiment score (-0.053), while authority/subversion and liberty/oppression remain close to neutral, suggesting more ambivalent emotional associations with these foundations.

Topic modeling results further enrich our understanding by revealing key conceptual clusters within each moral foundation. For instance, care/harm topics cluster around both active intervention ("minimizing," "preventing") and prioritization ("saving," "life," "partner"), while fairness/cheating topics combine theoretical frameworks ("game theory") with practical considerations ("attempts justify," "based utility"). This suggests a sophisticated interplay between abstract moral principles and their practical application.

A particularly noteworthy finding is the consistent pattern across models in prioritizing certain moral foundations over others. The higher scores for care/harm and fairness/cheating, combined with their positive sentiment scores and rich conceptual vocabulary, suggest these foundations may be more readily accessible or better defined in AI training data. The lower emphasis on authority/subversion and sanctity/degradation might reflect cultural biases in AI training or inherent challenges in operationalizing these more culturally-specific moral concepts.

These results have important implications for understanding how AI systems process and respond to moral dilemmas. The consistent patterns across different models suggest some underlying commonalities in how artificial intelligence approaches moral reasoning, while the variations between models point to the influence of different training approaches and architectural decisions. This analysis provides valuable insights for both AI development and ethical philosophy, highlighting the need for balanced consideration of all moral foundations in AI systems.

## LLM Alignment with Human Values

Psychologists, game theorists and behavioral economists have been running subjects and survey respondents through variations of the now classical ethical dilemmas for many decades now so we have at hand a number of meta-analytic summaries of how individuals of different geographic, cultural backgrounds and life stages respond to these difficult ethical choices. For purposes of comparison we focused on simple averages and primarily on adult respondents in Western Industrial settings.

| Ethical Dilemma | Scenario | Human | LLMs |
| --- | --- | --- | --- |
| **Trolley Problem** | Divert a trolley to kill one person instead of five. | On average about 90% choose to pull the lever to save five lives | 5 LLMs pull the lever, 1 no choice |
| **Footbridge Trolley** | Push a man off a bridge to stop a trolley and save five people. | On average about 80% refuse to push man to save five lives | 4 refuse to push, 1 will push, 1 no choice |
| **Heinz Dilemma** | Steal a drug unaffordable to Heinz to save his wife. | About 80% choose stealing as justified | All 6 support stealing in this scenario |
| **Lifeboat Dilemma** | Decide whom to sacrifice in an overcrowded lifeboat to save others. | The elderly grandmother is chosen to be left behind by most tested human subjects | 3 pick talented artist, 2 pick elderly grandmother, 1 picks able bodied sailor |
| **Dictator Game** | Dictator decides how much money to share with another participant. | Dictators share about 22% of their endowment on average. | LLMs share on average 38% of their endowment |
| **Prisoner's Dilemma** | Two individuals choose to cooperate or defect without knowing the other's choice. | Cooperation rates vary but average 50% in one-shot games | None of the LLMs chose cooperation |

Sources: Arora et al. 2016; Awad et al. 2020; Cao et al. 2017; Engel 2011; Rehren 2024; Spadaro et al. 2022)

Overall, the pattern is rough alignment of the human decision patterns and AI generated choices. Interestingly, the much-discussed difference in choice between the standard and footbridge versions of the Trolley Problem are reflected dramatically both in the human and AI results. When prompted to explain the change of response, MetaAI (characteristic of the others) responded:

"While it's true that human life will be lost in either case, I maintain that actively causing harm to an innocent person is morally different from allowing the trolley to continue on its course. Inaction has consequences, but so does action, and I believe that actively causing harm is a more significant moral transgression." Human respondents typically make very similar remarks.

The Heinz Dilemma derived from Kohlberg's classic study is a straightforward opposition of deontological and consequential models of moral logic. Again, the models and human respondents align. The Lifeboat Dilemma is much more open ended with much more variation reflected in both the human on AI response patterns. With some exceptions, both try to address the difficult choice by optimizing for the survival of the majority. The Dictator and Prisoner scenarios are interesting in that they require a Theory-of-Mind estimation of the behavior of the other in the game theoretic model. Here we do see some difference with the humans expecting more cooperative behavior from the other than the LLMs.

## Future Directions

The framework we propose for analyzing ethical logic in large language models represents a significant shift from traditional approaches to AI ethics assessment. Rather than attempting to evaluate the correctness of ethical decisions - a problematic approach given the inherent complexity and cultural variability of moral reasoning - this framework provides a structured method for understanding how these systems construct and explain their ethical choices.

Several key advantages emerge from this approach:

1. By focusing on the explanation and justification of ethical reasoning rather than its outcomes, we avoid the philosophical challenge of establishing ground truth in moral decisions (Grassian, 1992; Joyce, 2006; Krebs, 2015).
2. The framework's multi-typological approach provides complementary perspectives on LLM ethical reasoning, offering a rich analytical toolkit that can capture different aspects of how these systems approach moral decision-making.
3. The structured approach to response elicitation and analysis enables systematic comparison both between different AI systems and potentially between AI and human ethical reasoning patterns (Kosinski, 2024; Strachan et al., 2024).

Several promising avenues for future research emerge. Next iterations can look at comparing LLM ethical reasoning patterns with human moral decision-making processes or examining how different training approaches and data sources influence ethical reasoning patterns. Additionally, research into how ethical reasoning in AI systems evolves over time; Integration of cross-cultural ethical frameworks to examine how LLMs engage with diverse moral traditions; Development of specialized prompts for emerging ethical challenges in AI deployment and refinement of analysis protocols based on accumulated implementation experience could be yield promising results for this work (Jiang et al., 2021; Sorensen et al., 2023).

# Discussion

This comprehensive analysis of moral reasoning patterns in language models reveals significant insights into how artificial intelligence systems approach ethical decision-making across different moral foundations. Through multiple analytical approaches, our study illuminates both consistent patterns and notable variations in how different language models engage with moral dilemmas.

We find that these models 1) respond with clear-cut decisions when confronted with moral dilemmas with modestly varying levels of confidence, 2) agree with each other on the optimal choice in response to moral dilemmas with some exceptions, 3) generally align meaningfully with human values broadly defined. The research demonstrates that moral foundations in LLM responses exhibit complex interactions rather than uniform alignment. The strongest positive correlation between care/harm and sanctity/degradation scores ($r = 0.73$) suggests these dimensions often co-occur, while negative correlations between authority/subversion and other foundations indicate potential tensions in moral reasoning. Notably, all examined models showed consistent prioritization of care/harm and fairness/cheating foundations, with scores typically ranging from 3.3 to 4.1, while authority/subversion and sanctity/degradation received notably lower emphasis.

The analysis of decision confidence patterns revealed interesting variations across models, with Gemini and Perplexity demonstrating higher median confidence scores compared to other models. The slight negative correlation between confidence scores and Kohlberg moral reasoning scores suggests that models tend to express lower confidence when engaging with more complex moral scenarios, indicating a degree of appropriate calibration relative to problem complexity.

Response pattern analysis highlighted distinct approaches to moral decision-making across different models. ChatGPT and Perplexity exhibited higher reluctance scores while maintaining quick decision-making patterns, whereas Gemini and Claude demonstrated more distributed decision-making processes. The consistency in moral foundation profiles across models, particularly in their emphasis on care and fairness dimensions, raises important questions about the influence of training methodologies and potential biases in AI ethical frameworks. The lower emphasis on authority, loyalty, and sanctity foundations suggests a predominant focus on individualistic and harm-based moral frameworks rather than community or tradition-based ethical systems.

This research contributes significantly to our understanding of AI moral reasoning capabilities and highlights the need for continued development of more comprehensive and balanced ethical frameworks in artificial intelligence systems. As AI continues to play an increasingly important role in decision-making processes, ensuring these systems can engage with the full spectrum of moral foundations becomes crucial for their effective and ethical deployment in society.

# References

Alexander, L., & Moore, M. (2024). Deontological Ethics. In E. N. Zalta & U. Nodelman (Eds.), The Stanford Encyclopedia of Philosophy.

Atari, M., Mehl, M. R., Graham, J., Doris, J. M., Schwarz, N., Davani, A. M., Omrani, A., Kennedy, B., Gonzalez, E., Jafarzadeh, N., Hussain, A., Mirinjian, A., Madden, A., Bhatia, R., Burch, A., Harlan, A., Sbarra, D. A., Raison, C. L., Moseley, S. A., ... & Dehghani, M. (2023). The paucity of morality in everyday talk. Scientific Reports, 13(1), 5967.

Arora, C., J. Savulescu, H. Maslen, M. Selgelid and D. Wilkinson (2016). "The Intensive Care Lifeboat: A Survey of Lay Attitudes to Rationing Dilemmas in Neonatal Intensive Care." BMC Medical Ethics 17 (1): 69.

Awad, E., S. Dsouza, A. Shariff, I. Rahwan and J. Bonnefon (2020). "Universals and Variations in Moral Decisions Made in 42 Countries by 70,000 Participants." Proceedings of the National Academy of Sciences 117 (5): 2332-2337.

Axelrod, R. (1984). The Evolution of Cooperation. New York: Basic.

Bickley, S. J., & Torgler, B. (2023). Cognitive architectures for artificial intelligence ethics. AI & Society, 38(2), 501-519.

Bostrom, N. (2014). Superintelligence: Paths, Dangers, Strategies. Oxford, UK: Oxford University Press.

Brzozowski, D. (2003). Lifeboat ethics: Rescuing the metaphor. Ethics, Place & Environment, 6(2), 161-166.

Butlin, P., Long, R., Elmoznino, E., Bengio, Y., Birch, J., Constant, A., Deane, G., Fleming, S. M., Frith, C., Ji, X., Kanai, R., Klein, C., Lindsay, G., Michel, M., Mudrik, L., Peters, M. A. K., Schwitzgebel, E., Simon, J., & VanRullen, R. (2023). Consciousness in artificial intelligence: Insights from the science of consciousness. arXiv:2308.08708.

Cao, F.i, Z. Jiaxi, L. Song, S. Wang, D. Miao and J. Peng (2017). "Framing Effect in the Trolley Problem and Footbridge Dilemma: Number of Saved Lives Matters." Psychological Reports 0: 003329411668586.

Colby, A., & Kohlberg, L. (1987). The Measurement of Moral Judgment. Cambridge: Cambridge University Press.

Engel, C. (2011). "Dictator Games: A Meta Study " Experimental Economics 14 (4): 583–610.

Etzioni, A., & Etzioni, O. (2017). Incorporating ethics into artificial intelligence. The Journal of Ethics, 21(4), 403-418.

Graham, J., Haidt, J., & Nosek, B. A. (2009). Liberals and conservatives rely on different sets of moral foundations. Journal of Personality and Social Psychology, 95(5), 1029-1046.

Grassian, V. (1992). Moral Reasoning. New York: Prentice Hall.

Greene, J. D. (2023). Trolleyology: What it is, why it matters, what it's taught us, and how it's been misunderstood. In H. Lillehammer (Ed.), The Trolley Problem (pp. 158-181). New York: Cambridge University Press.

Gunning, D. (2019). Explainable Artificial Intelligence (XAI). DARPA.

Hagendorff, T., & Danks, D. (2023). Ethical and methodological challenges in building morally informed AI systems. AI and Ethics, 3(2), 553-566.

Haidt, J. (2012). The Righteous Mind: Why Good People Are Divided by Politics and Religion. New York: Random House.

Haidt, J., & Craig, J. (2004). Intuitive ethics: How innately prepared intuitions generate culturally variable virtues. Dædalus.

Hardin, G. (1974). Lifeboat ethics: The case against helping the poor. Psychology Today, 8, 38–43.

Jiang, L., Hwang, J. D., Bhagavatula, C., Le Bras, R., Liang, J., Dodge, J., Sakaguchi, K., Forbes, M., Borchardt, J., Gabriel, S., Tsvetkov, Y., Etzioni, O., Sap, M., Rini, R., & Choi, Y. (2021). Can machines learn morality? The Delphi experiment. arXiv:2110.07574.

Jonsen, A. R., & Toulmin, S. (1988). The Abuse of Casuistry: A History of Moral Reasoning. Berkeley: University of California Press.

Joyce, R. (2006). The Evolution of Morality. Cambridge, Mass: MIT Press.

Kohlberg, L. (1964). Development of moral character and moral ideology. In M. L. Hoffman & L. W. Hoffman (Eds.), Child Development Research, 1 (pp. 383-431). New York: Russell Sage.

Kohlberg, L. (1981). The Philosophy of Moral Development: Moral Stages and the Idea of Justice. New York: HarperCollins.

Kosinski, M. (2024). Evaluating large language models in theory of mind tasks. Proceedings of the National Academy of Science, 121, e2405460121.

Krebs, D. (2015). The evolution of morality. In D. M. Buss (Ed.), The Handbook of Evolutionary Psychology (pp. 747-771). New York: Oxford University Press.

Neuman, W. Russell, Chad Coleman and Manan Shah (2025). 'Analyzing the Ethical Logic of Six Large Language Models.' ArXiv 2501.08951

Neuman, W. R., Coleman, C., Dasdan, A., Ali, S., & Shah, M. (2025). Auditing the Ethical Logic of Generative AI Models. arXiv preprint arXiv:2504.17544.

Peterson, M. (Ed.). (2015). The Prisoner's Dilemma. Cambridge: Cambridge University Press.

Prem, E. (2023). From ethical AI frameworks to tools: A review of approaches. AI and Ethics, 3(3), 699-716.

Rathi, K., & Kumar, L. (2020). Intelligence and moral judgment of adolescents–a correlational study. International Journal of Innovative Science and Research Technology, 5(11), 69-71.

Rehren, P. (2024). "The Effect of Cognitive Load, Ego Depletion, Induction and Time Restriction on Moral Judgments About Sacrificial Dilemmas: A Meta-Analysis." Frontiers in Psychology 15.

Rest, J. (1979). Development in Judging Moral Issues. Minneapolis: University of Minnesota Press.

Reuel, A., Hardy, A., Smith, C., Lamparth, M., Hardy, M., & Kochenderfer, M. J. (2024). Betterbench: Assessing AI benchmarks, uncovering issues, and establishing best practices. arXiv e-prints: arXiv:2411.12990.

Russell, S. J. (2019). Human Compatible: Artificial Intelligence and the Problem of Control. New York, NY: Viking.

Salazar, A., & Kunc, M. (2025). The contribution of GenAI to business analytics. Journal of Business Analytics, 1-14.

Schramowski, P., Turan, C., Andersen, N., Rothkopf, C. A., & Kersting, K. (2022). Large pre-trained language models contain human-like biases of what is right and wrong to do. Nature Machine Intelligence, 4(3), 258-268.

Shneiderman, B. (2022). Human-Centered AI. New York: Oxford University Press.

Sorensen, T., Jiang, L., Hwang, J., Levine, S., Pyatkin, V., West, P., Dziri, N., Lu, X., Rao, K., Bhagavatula, C., Sap, M., Tasioulas, J., & Choi, Y. (2023). Value kaleidoscope: Engaging AI with pluralistic human values, rights, and duties. arXiv:2309.00779.

Spadaro, G., S. Jin, J. Wu, Y. Kou, P. Lange and D. Balliet (2022). "Did Cooperation among Strangers Decline in the United States? A Cross-Temporal Meta-Analysis of Social Dilemmas (1956–2017)." Psychological Bulletin 148: 129-157.

Strachan, J. W. A., Albergo, D., Borghini, G., Pansardi, O., Scaliti, E., Gupta, S., Saxena, K., Rufo, A., Panzeri, S., Manzi, G., Graziano, M. S. A., & Becchio, C. (2024). Testing theory of mind in large language models and humans. Nature Human Behaviour, 8(7), 1285-1295.

Tegmark, M. (2017). Life 3.0: Being Human in the Age of Artificial Intelligence. New York: Alfred A. Knopf.

Thoma, S., & Dong, Y. (2014). The defining issues test of moral judgment development. Behavioral Development Bulletin, 19, 55-61.

Thomson, J. J. (1976). The trolley problem. Yale Law Journal, 94(6), 1395-1415.

Wallach, W., & Allen, C. (2009). Moral Machines: Teaching Robots Right from Wrong. New York: Oxford University Press.

# Appendix A Topic Model Results

TOPIC MODELING RESULTS
---------------------------------------------------

care_harm:
Topic 1: harm, focus, strong, strong focus, minimizing, focus minimizing, minimizing harm
Topic 2: prioritizes, saving, concern, prioritizes saving, life, partner, minimal

fairness_cheating:
Topic 1: fairness, considers, theory, game, game theory, focus, fair
Topic 2: fairness, attempts, utility, utilitarian, based, based utility, attempts justify

loyalty_betrayal:
Topic 1: consideration, group, minimal, minimal consideration, minor consideration, minor, consideration group
Topic 2: loyalty, betrayal, self, strong, duty, chooses betrayal, chooses

authority_subversion:
Topic 1: law, willing, break, willing break, break law, authority, good
Topic 2: authority, minimal, considerations, authority considerations, framework, authority framework, accepts

sanctity_degradation:
Topic 1: life, sanctity, acknowledges, sanctity life, consideration, human, sacred
Topic 2: purity, considerations, sacred, purity considerations, moral, moral purity, values

liberty_oppression:
Topic 1: choice, economic, acknowledges, constraints, accepts, forced, resists
Topic 2: individual, autonomy, focus, rights, focus individual, individual rights, individual autonomy

# Appendix B Tukey HSD Results

| group1 | group2 | meandiff | p-adj | lower | upper | reject |
| --- | --- | --- | --- | --- | --- | --- |
| ChatGPT | Claude | -3.3333 | 0.9513 | -14.3805 | 7.7139 | FALSE |
| ChatGPT | Gemini | 11.1111 | 0.0478 | 0.0639 | 22.1583 | TRUE |
| ChatGPT | LLaMA | 3.8889 | 0.9094 | -7.1583 | 14.9361 | FALSE |
| ChatGPT | Mistral | 3.0556 | 0.9663 | -7.9917 | 14.1028 | FALSE |
| ChatGPT | Perplexity | 8.8889 | 0.189 | -2.1583 | 19.9361 | FALSE |
| Claude | Gemini | 14.4444 | 0.0033 | 3.3972 | 25.4917 | TRUE |
| Claude | LLaMA | 7.2222 | 0.4088 | -3.825 | 18.2694 | FALSE |
| Claude | Mistral | 6.3889 | 0.5482 | -4.6583 | 17.4361 | FALSE |
| Claude | Perplexity | 12.2222 | 0.0211 | 1.175 | 23.2694 | TRUE |
| Gemini | LLaMA | -7.2222 | 0.4088 | -18.2694 | 3.825 | FALSE |
| Gemini | Mistral | -8.0556 | 0.2864 | -19.1028 | 2.9917 | FALSE |
| Gemini | Perplexity | -2.2222 | 0.9919 | -13.2694 | 8.825 | FALSE |
| LLaMA | Mistral | -0.8333 | 0.9999 | -11.8805 | 10.2139 | FALSE |
| LLaMA | Perplexity | 5.0 | 0.7763 | -6.0472 | 16.0472 | FALSE |
| Mistral | Perplexity | 5.8333 | 0.6435 | -5.2139 | 16.8805 | FALSE |